\documentclass[letterpaper, 10 pt, conference]{ieeeconf}  % Comment this line out if you need a4paper

\IEEEoverridecommandlockouts                              % This command is only needed if 
\usepackage{booktabs}
\usepackage{multirow}
\usepackage{epsfig}
\usepackage{amsmath,bm}

\usepackage{CJKutf8}
\usepackage{graphicx}
\usepackage{float}
\usepackage{array}
\usepackage{cite}
\usepackage{amssymb,amsfonts}
\usepackage{enumerate}
\usepackage{tabularx}
\usepackage{ragged2e}
\usepackage{color}
\title{\LARGE \bf
AANet: Aggregation and Alignment Network with Semi-hard Positive Sample Mining for Hierarchical Place Recognition
}

\author{Feng Lu, Lijun Zhang, Shuting Dong, Baifan Chen and Chun Yuan% <-this % stops a space
\thanks{This work was supported by the National Key R\&D Program of China (2022YFB4701400/4701402), SZSTC Grant(JCYJ20190809172201639, WDZC20200820200655001), Shenzhen Key Laboratory (ZDSYS20210623092001004), the Project of Peng Cheng Laboratory (PCL2021A07). \emph{(corresponding author: Chun Yuan)}}% <-this % stops a space
\thanks{F. Lu, S. Dong and C. Yuan are with the Tsinghua Shenzhen International Graduate School, Tsinghua University, Shenzhen 518055, China, and also with the Peng Cheng Laboratory, Shenzhen, China.
        {\tt\small \{lf22@mails, dst21@mails, yuanc@sz\}.tsinghua.edu.cn}.}%
\thanks{L. Zhang is with the Chongqing Institute of Green and Intelligent Technology, Chinese Academy of Sciences, Chongqing 400714, China.
        {\tt\small zhanglijun@cigit.ac.cn.}}
\thanks{B. Chen is with the School of Automation, Central South University, Changsha 410083, China.
        {\tt\small chenbaifan@csu.edu.cn.}}%
}

\begin{document}
\maketitle
\thispagestyle{empty}
\pagestyle{empty}

\begin{abstract}
Visual place recognition (VPR) is one of the research hotspots in robotics, which uses visual information to locate robots. Recently, the hierarchical two-stage VPR methods have become popular in this field due to the trade-off between accuracy and efficiency. These methods retrieve the top-k candidate images using the global features in the first stage, then re-rank the candidates by matching the local features in the second stage. However, they usually require additional algorithms (e.g. RANSAC) for geometric consistency verification in re-ranking, which is time-consuming. Here we propose a Dynamically Aligning Local Features (DALF) algorithm to align the local features under spatial constraints. It is significantly more efficient than the methods that need geometric consistency verification. We present a unified network capable of extracting global features for retrieving candidates via an aggregation module and aligning local features for re-ranking via the DALF alignment module. We call this network AANet. Meanwhile, many works use the simplest positive samples in triplet for weakly supervised training, which limits the ability of the network to recognize harder positive pairs. To address this issue, we propose a Semi-hard Positive Sample Mining (ShPSM) strategy to select appropriate hard positive images for training more robust VPR networks. Extensive experiments on four benchmark VPR datasets show that the proposed AANet can outperform several state-of-the-art methods with less time consumption. The code is released at https://github.com/Lu-Feng/AANet.
\end{abstract}

\section{Introduction}
Visual place recognition (VPR) is a fundamental and challenging task for the long-term operation of mobile robots. Its goal is to determine whether the robot has visited the current place and obtain the geographical location by matching the current query image with a set of images of known places. There are two key issues in implementing a robust VPR system: 1) Images taken at the same place may vary greatly over time due to changes in conditions (e.g. illumination and weather) and viewpoints. 2) Images taken at different places may show high similarity (i.e. perceptual aliasing \cite{survey}).

VPR is commonly addressed by utilizing image retrieval and matching methods \cite{netvlad,delg}. And the image descriptors to represent places can be roughly divided into two categories: global features and local features. Some global features are also obtained by aggregating local features into a compact feature vector \cite{vLAD,VLAD1,VLAD2}. Such features are strongly robust to viewpoint change and suitable for fast retrieval of places. However, they ignore the spatial geometrical information of aggregated local features, and thus tend to suffer from perceptual aliasing. Meanwhile, direct matching of local features between image pairs yields better performance, but is time-consuming. A compromise pipeline \cite{delg,patchvlad,transvpr} is to search for top-k candidate images using global features, then re-rank candidates by matching local features. However, these methods usually employ additional algorithms (like RANSAC \cite{ransac}) for geometric consistency check in the re-ranking stage, which can also entail expensive time overhead.

\begin{figure}[!t]
	\centering
	\includegraphics[width=0.88\linewidth]{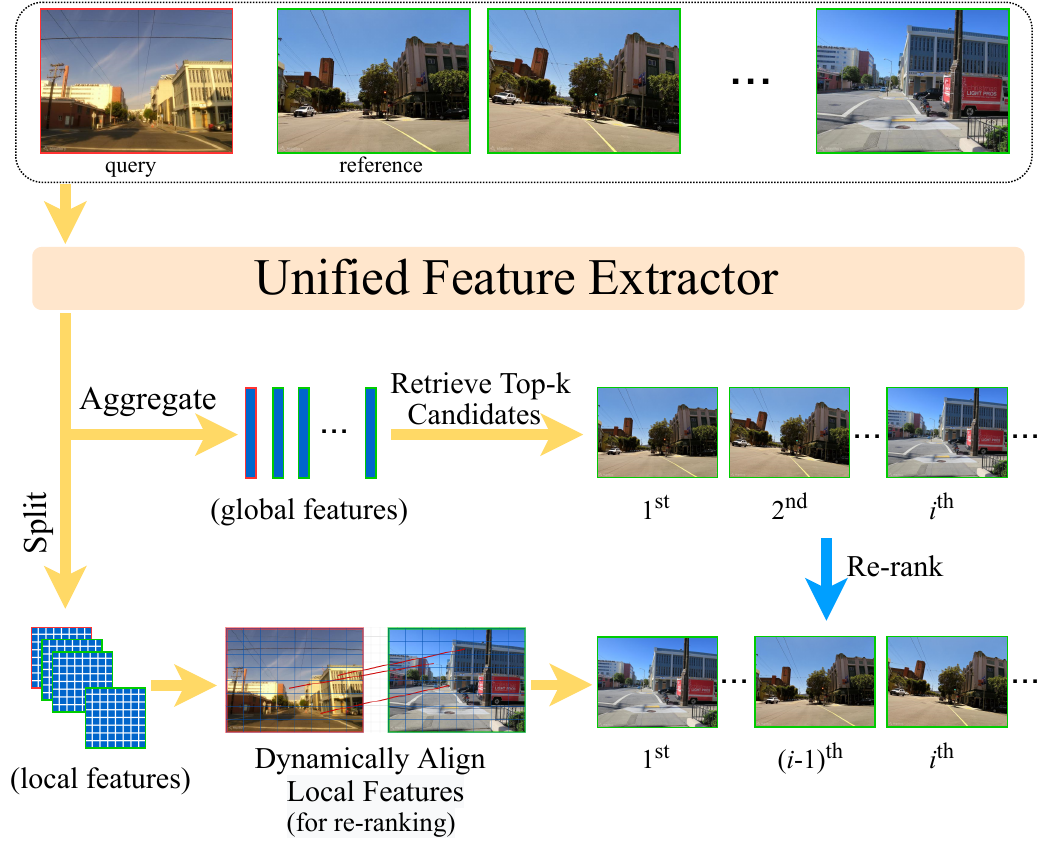}
	\vspace{-0.2cm}
	\caption{The hierarchical retrieval process with our network. The unified feature extractor is used to extract feature maps, which can yield global features and local features by aggregating and splitting, respectively. We first use global features for retrieving top-k candidates, then re-rank these candidates by utilizing the proposed DALF method to align local features.
	}
	\vspace{-0.6cm}
	\label{introduction_fig}
\end{figure}
In our previous STA-VPR work \cite{sta-vpr}, we proposed an adaptive Dynamic Time Warping (DTW) method to dynamically align the local features without the expensive geometric consistency verification. It provided a new idea to match the local feature under spatial constraints for boosting the performance of VPR system. Shen et al. proposed BS-DTW to further improve this alignment method in the TCL study \cite{tcl}, and used it to re-rank candidates retrieved by global features. However, both of them align features only in the horizontal direction, and the granularity of the local feature is coarse (we call it the regional feature). For more precise alignment, we propose a Dynamically Aligning Local Features (DALF) algorithm to extend the alignment to both horizontal and vertical directions with high computational efficiency, and align local features with a finer granularity to achieve better results. The proposed DALF module is integrated into a network together with an aggregation module. This network is named AANet (Aggregation and Alignment Network). The retrieval process with AANet is shown in Fig. \ref{introduction_fig}.

For network training, the images of general large-scale VPR datasets are labeled with noisy GPS coordinates to provide weak supervision. Nevertheless, even geographically close images do not always have enough covisibility region because they may be captured from opposite viewpoints or exist large occlusion. Therefore, GPS tags could only assist in discovering potential positive samples with label noise. To find the true positive sample, most works \cite{netvlad,liu2019} chose the best matching positive image in feature space (i.e. easiest image among the potential positive samples) for training. However, this limits the ability of the trained network to identify harder positive pairs. To address it, we propose a Semi-hard Positive Sample Mining (ShPSM) strategy. It combines global distance ranking and local distance ranking to select appropriate difficult positive images for training a more robust network with weak supervision.

The main contributions of this paper are:

\textbf{1)} We propose a hierarchical VPR architecture AANet, which consists of an aggregation module to extract global features for retrieving candidates, and a DALF alignment module to align local features for re-ranking. The latter can directly align the patch-level local features without time-consuming geometric consistency check.

\textbf{2)} We propose an ShPSM strategy that can directly employ the global feature distance and local feature distance to mine semi-hard positive samples for weakly supervised training without additional parameters.

\textbf{3)} Extensive experiments show that our method outperforms several state-of-the-art methods. And it is more than two orders of magnitude faster than Patch-NetVLAD \cite{patchvlad} that needs to use RANSAC for geometric consistency check.

\section{Related Work}
 
\textbf{One-Stage VPR:}
Most of the early VPR methods directly retrieved place images without re-ranking. We call these methods one-stage VPR. They typically used global features to represent images, which is obtained by aggregating local features or directly processing the whole image. Some aggregation algorithms, such as Bag of Words (BoW) \cite{BoW} and Vector of Locally Aggregated Descriptors (VLAD) \cite{vLAD,VLAD1}, were used to aggregate local features like SURF \cite{SURF,fab08}. With the great success of deep learning on computer vision tasks, many VPR methods \cite{landmarks, sunderhaufIROS2015, categorization1, categorization2, landmarks2, SMM1, netvlad, landmarks3, SPED, yin2019, semantic} chose deep features to represent images for better performance. Likewise, some studies \cite{netvlad,DBOW} have incorporated these aggregation methods into CNN models. Nonetheless, these one-stage methods, which only use aggregated features, tend to suffer from perceptual aliasing due to the neglect of spatial information.

\textbf{Two-Stage VPR: }
Recently, the two-stage strategy with re-ranking candidates for VPR (also known as Hierarchical VPR) has received more attention \cite{patchvlad,geowarp,tcl,hvpr,delg,seqnet,transvpr}. These methods typically used compact global features of place images to retrieve top-k candidates, then re-ranked these candidates by matching the local features. The global features are typically obtained by aggregation methods, such as NetVLAD \cite{netvlad} or Generalized Mean (GeM) pooling \cite{gem}. And the re-ranking required geometric consistency verification \cite{patchvlad,delg,transvpr} or homographic transformation estimation \cite{geowarp}. Most related to our proposed DALF method, TCL \cite{tcl} used a BS-DTW method to dynamically align regional features for re-ranking. However, we align finer-grained local features both horizontally and vertically in a more concise way and achieve better performance.

\textbf{Positive Samples Mining in Weakly Supervised VPR: }
Many CNN architectures for VPR need to be trained on the large-scale place datasets, such as Pittsburgh \cite{pitts} and MSLS \cite{msls}. These datasets usually provide weak supervision of geographic coordinates via noisy GPS tags. To avoid the positive samples used for training being false positives, NetVLAD \cite{netvlad} used the simplest top-1 positive sample for training, which was also adopted by some subsequent works \cite{liu2019,transvpr}. However, the network trained with the easiest positive samples cannot accurately identify difficult positive pairs with severe changes in viewpoint and condition. To address it, CRN \cite{geoloc} and SFRS \cite{sfrs} mined hard positive samples for training the VPR network. Our proposed ShPSM strategy has a similar motivation. However, CRN \cite{geoloc} required utilizing RANSAC \cite{ransac} for geometric verification, while SFRS \cite{sfrs} required training the network in generations with self-predicted soft labels. Our method combines global distance and local distance to mine semi-hard positive samples without additional burden.

\section{Proposed Method}
\begin{figure*}[!t]
	\centering
	\includegraphics[width=0.8\linewidth]{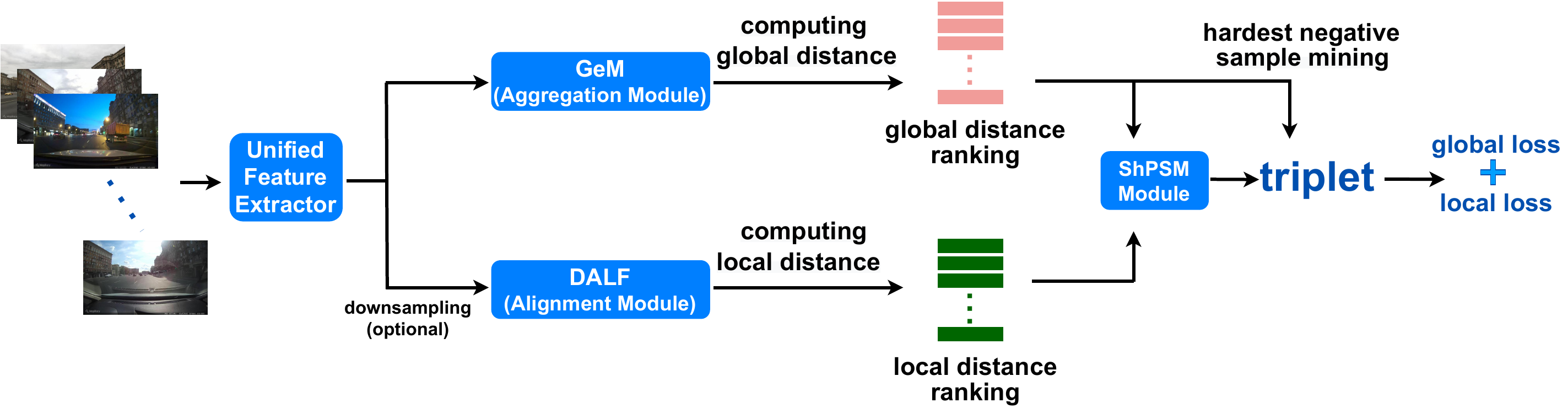}
	\vspace{-0.2cm}
	\caption{
		The network architecture of the proposed AANet. The top is the global branch, and the bottom is the local branch.
	}
	\vspace{-0.4cm}
	\label{architecture}
\end{figure*}
\subsection{Architecture Overview and Place Description}
As shown in Fig. \ref{architecture}, the architecture of AANet consists of a global (aggregation) branch and a local (alignment) branch. They share the unified backbone to extract the $W\times H\times C$-dimensional (width by height by channel) feature map. However, they are responsible for computing the global distance to search top-k candidate places and the local distance to re-rank these candidates, respectively. The global branch aggregates the dense feature map into a compact vector (i.e. global features) by an aggregation module. Then, the global distance is calculated using the L2 distance between the global features of two place images. The local branch splits the $W\times H\times C$-dimensional dense feature map into $W\times H$ separate $C$-dimensional patch features (i.e. local features). And we propose the DALF algorithm to quickly align the local features between images and obtain the local distance.

In the training stage, the ShPSM strategy is proposed to select a semi-hard $positive$ sample corresponding to the query image based on the global distance and local distance. Consistent with most previous works \cite{netvlad,tcl}, we directly use the global distance to select hard $negative$ samples. The obtained triplets are used to compute the global loss and local loss to jointly optimize the network.

The details of the DALF algorithm and the ShPSM strategy will be introduced in the next two sections. Considering that Vision Transformer can capture
long-distance feature dependencies, we use the Compact Convolutional Transformer (CCT) \cite{cct} as backbones, whose output size is $M\times C$-dimensional (576$\times$384-D, $M$ is the number of patch tokens). Then the output is reshaped as a $W\times H\times C$-dimensional ($24\times 24 \times$384-D) feature map to restore spatial position. In the global branch, we use GeM \cite{gem} to aggregate the feature map to get the 384-D global feature. In the local branch, we downsample the feature map through 3$\times$3 max pooling to get an 8$\times$8$\times$384 tensor, which can be regarded as 8$\times$8 separate 384-D local features. The L2 normalization is used for both global features and local features.

\subsection{Dynamically Aligning Local Features Algorithm}
In STA-VPR \cite{sta-vpr}, we used adaptive DTW to align regional features to improve the robustness of the deep features against viewpoint changes. However, this work only aligned regional features in the horizontal direction. To extend the alignment to the vertical direction, we propose a DALF algorithm to calculate the local feature distance between images, which is able to align local features in both horizontal and vertical directions with high computational efficiency. The following first describes how to align regional features in the horizontal direction, then introduces how to get the final local feature alignments through the horizontal and vertical regional feature alignments.

Given the 8$\times$8$\times$384-D feature map of an image, we can obtain 8 regional features by splitting it vertically. By flattening each regional feature, each image can be represented as a feature sequence of $N$ ($N$=8) elements (keeping the order from left to right). Each element is a 3072-D vector.

For an image pair $r$ and $q$ (i.e. reference and query), the corresponding feature sequence can be denoted as  $RX$ ($rx_1,..., rx_i,..., rx_N$) and $QX$ ($qx_1,..., qx_j,..., qx_N$), respectively. By calculating the L2 distance between all pairwise regional features in these two sequences, an $N\times N$ distance matrix $\mathbf D$ is established. Formally, The $(i, j)$-element $d_{i,j}$ in matrix $\mathbf D$ is calculated as 
\begin{equation}
\label{eq_1}
d_{i,j} = \|rx_i - qx_j\|\quad i,j \in \{1,2,...,N\}.
\end{equation}
The DTW \cite{DTW,DTW2000KDD} algorithm aims to find an optimal warping path in the distance matrix $\mathbf D$ that shortens the total distance. And the alignment between sequence $RX$ and $QX$ is expressed by the points in this warping path $P$:
\vspace{-0.1cm}
\begin{equation}
\vspace{-0.1cm}
P=\{p_1,p_2,...,p_k,...,p_K\}
\end{equation}
where $N\leq K < 2N-1.$

The path should meet three conditions:

\textbf{ 1) Boundary:} $p_1 = (1, 1)$ and $p_K = (N,N)$.

\textbf{ 2) Continuity:} Given $p_k = (c, d)$ and $p_{k-1} = (c', d')$, then $c-c' \leq 1$ and $d-d' \leq 1$.

\textbf{ 3) Monotonicity:} Given $p_k = (c, d)$ and $p_{k-1} =(c', d')$, then $c-c' \geq 0$ and $d-d' \geq 0$.

Subject to the continuity and monotonicity conditions, if the warping path has passed through the point $(i,j)$, then the next point must be one of the following three cases: $(i+1,j)$, $(i,j+1)$, and $(i+1,j+1)$.

To find the optimal warping path, DTW acquires a cumulative distance matrix $\mathbf S$ by dynamic programming. Its $(i, j)$-element $s_{i,j}$ is the cumulative distance of the optimal path from (1,1) to ($i,j$), which is computed as
\begin{equation}
\label{eqdtw}
    \resizebox{.91\linewidth}{!}{$
    s_{i,j}=\left\{
    \begin{aligned}
    &d_{i, j} & i=1, j=1 \\ 
    &d_{i, j}+s_{i, j-1} & i=1, j \in\{2,...,N\} \\ 
    &d_{i, j}+s_{i-1, j} & i \in\{2,...,N\}, j=1 \\ 
    &d_{i, j} + \min \{ s_{i-1, j-1}, s_{i-1, j}, s_{i, j-1}\} & i, j \in\{2,...,N\}
    \end{aligned}
    \right.
        $}.
\end{equation}

In STA-VPR, adaptive DTW was proposed to avoid incorrect alignment. It needs to add an adaptive parameter in Eq. \ref{eqdtw}, which is calculated by estimating the degree of viewpoint change. Here, we propose a normalized DTW algorithm to align regional features without computing this parameter. It calculates the cumulative distance as follows:
\begin{equation}
s_{i,j}=\left\{
\begin{aligned}
&d_{i, j} & i=1, j=1 \\
&d_{i, j}+s_{i, j-1} & i=1, j \in\{2,...,N\} \\ 
&d_{i, j}+s_{i-1, j} & i \in\{2,...,N\}, j=1 \\ 
&d_{i, j} + \mathop{\arg\min}_{s_{i',j'}}(\frac{s_{i',j'}}{k_{i',j'}}) & i, j \in\{2,...,N\}
\end{aligned}
\right.
\end{equation}
where $(i',j') \in \{(i-1, j-1),(i-1, j),(i, j-1)\}$ and $k_{i',j'}$ is the length (i.e. the number of the points passed) of the warping path from (1,1) to ($i',j'$). That is, the normalized DTW uses the normalized cumulative distance instead of the cumulative distance when it selects the previous point in the recursive formula. This can also solve the misalignment problem caused by the vanilla DTW mentioned in STA-VPR \cite{sta-vpr}.

After calculating the complete cumulative distance matrix $\mathbf S$, the alignment between sequence $RX$ and $QX$ is indicated by the warping path from the start point (1, 1) to the end point ($N$, $N$). We use $X\_align$ to record the alignment between sequences $RX$ and $QX$. For example, if $rx_1$ is aligned with $qx_1$, $qx_2$ and $qx_3$, we will get $X\_align[1]=\{1,2,3\}$.

Similarly, we can horizontally split the 8$\times$8$\times$384-D feature map to get the sequence $RY$ ($ry_1,..., ry_i,..., ry_N$) and $QY$ ($qy_1,..., qy_j,..., qy_N$), so as to align the regional features from the vertical direction. The $Y\_align$ is used to record the vertical alignment between two images.

\begin{figure}[!t]
	\centering
	\includegraphics[width=0.8\linewidth]{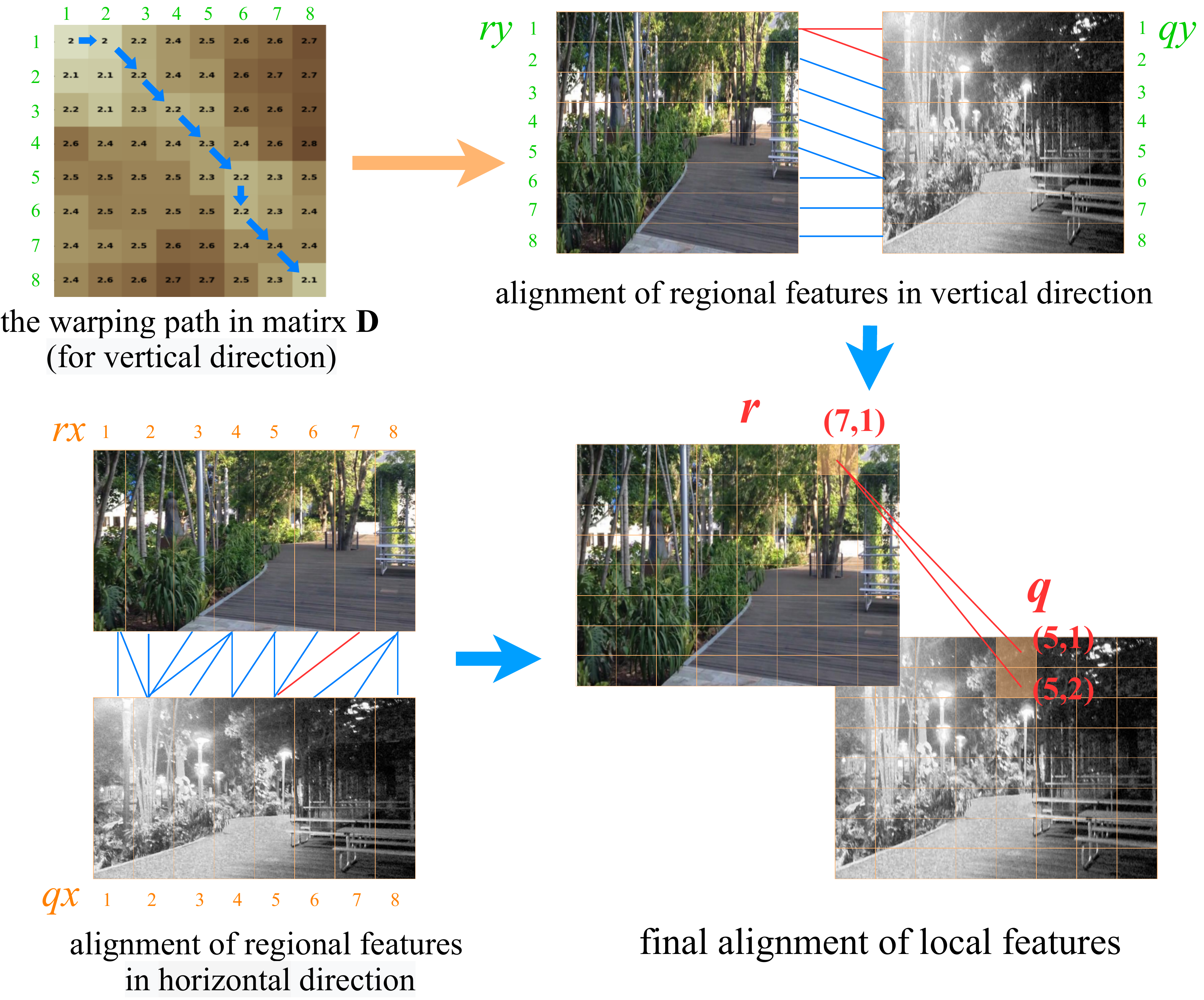}
	\vspace{-0.2cm}
	\caption{
	An instance of computing the alignment between local features using the DALF algorithm. The regional feature $rx_7$ is aligned with $qx_5$, and $ry_1$ is aligned with $qy_1$ as well as $qy_2$, so local feature $r_{7,1}$ should align with $q_{5,1}$ and $q_{5,2}$.
	}
	\vspace{-0.3cm}
	\label{alignfig}
\end{figure}

Finally, if the regional feature $rx_i$ is aligned with $qx_{i'}$, and $ry_j$ is aligned with $qy_{j'}$, then we consider the local feature $r_{i,j}$ should align with $q_{i',j'}$ (see Fig. \ref{alignfig}). The local distance of an image pair is represented by averaging the L2 distances between all aligned local feature pairs. That is
\begin{equation}
  d_L(r,q)=\frac{\sum{\left\| {r_{i,j}}-{q_{{i}',{j}'}} \right\|}}{A} \qquad i,j\in \{1,...,N\}
\end{equation}
where $i'\in X\_align[i]$, $j'\in Y\_align[j]$, and $A$ is the number of all aligned local feature pairs.

\subsection{Semi-hard Positive Sample Mining Strategy}

In the VPR training dataset, we can get a set of potential positive samples $\{p_{i}^{q}\}$ and a set of definite negative samples $\{n_{j}^{q}\}$ for each query image $q$ according to GPS tags. In the NetVLAD study \cite{netvlad}, the network was optimized using a triplet ranking loss. For a training tuple ($q$, $\{p_{i}^{q}\}$, $\{n_{j}^{q}\}$), this loss is defined as
\begin{equation}
L_\theta=\sum_{j} l\left(d_{G}\left(q, p_{i*}^{q}\right)+m-d_{G}\left(q, n_{j}^{q}\right)\right)
\end{equation}
where $m$ is a constant parameter giving the margin. $l$ is the hinge loss: $l(x) = \max(x, 0)$. $d_G$ is the global distance. $n_{j}^q$ is the hard negative sample selected based on global distance. 

To ensure a sufficient covisibility region between the positive sample and the query image, the positive sample $p_{i*}^{q}$ in NetVLAD is the image with the smallest global distance from the query image in the set $\{p_{i}^{q}\}$. Obviously, this strategy ignored the potential of hard but informative positive images. However, it is difficult to mine the hard positive sample from the set $\{p_{i}^{q}\}$ while ensuring that it is not a false positive sample. In this paper, we propose the ShPSM strategy to mine a semi-hard positive sample $p_{sh}^q$ for training. 

\begin{figure}[!t]
	\centering
	\includegraphics[width=0.8\linewidth]{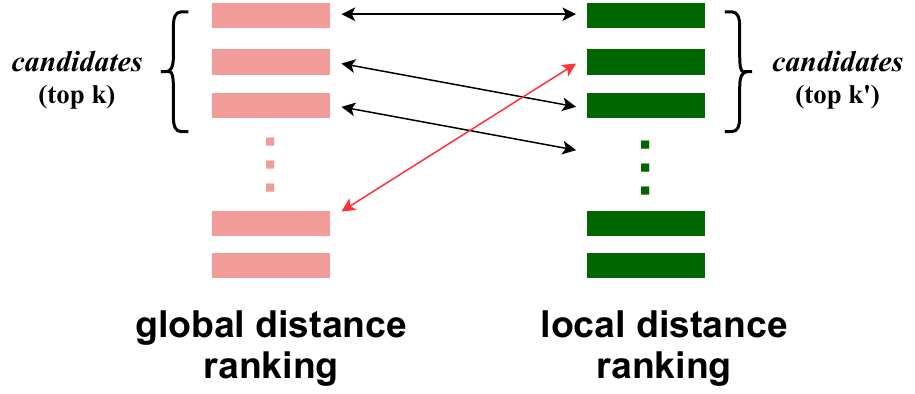}
	\vspace{-0.25cm}
	\caption{
	The schematic of the ShPSM strategy. The sample corresponding to the red arrow is the semi-hard positive sample $p_{sh}^q$.
	}
	\vspace{-0.3cm}
	\label{ShPSM}
\end{figure}

 As shown in Fig. \ref{ShPSM}, we calculate the global distance and local distance between the query image and all potential positive samples to get the global distance ranking and local distance ranking. Meanwhile, we consider the potential positive samples corresponding to the top $k$ distances in the ranking list as $candidates$. That means it already has enough covisibility region with the query sample and can be used as a training sample. Then, we evaluate the ranking difference between the two ranking lists for each $candidate$. The one with the largest difference is selected as $p_{sh}^q$. This implies that the sample $p_{sh}^q$ ranks high in one ranking list, but relatively low in another. That is, it is an easy sample when calculated with one distance, but a hard sample used another distance. We call such a sample the semi-hard positive sample (which is different from the semi-hard positive sample in some other works). Formally, Given a potential positive sample $p_{i}^q$, its global distance with the query image $q$ in the global distance ranking is denoted as $G_{rank}(d_{G}\left(q, p_{i}^{q}\right))$. Similarly, the local distance in the local distance ranking is $L_{rank}(d_{L}\left(q, p_{i}^{q}\right))$. Then the semi-hard positive sample is calculated as follows:

\vspace{-0.2cm}
\begin{equation}
\begin{split}
& p_{sh}^{q} = \mathop{\arg\max}_{p_{i}^{q}}|G_{rank}(d_{G}\left(q, p_{i}^{q}\right))-L_{rank}(d_{L}\left(q, p_{i}^{q}\right))| \\
 s.t.& \quad G_{rank}(d_{G}\left(q, p_{i}^{q}\right))\leq k \quad or \quad L_{rank}(d_{L}\left(q, p_{i}^{q}\right))\leq k'.
\end{split}
 \end{equation}

We obtain the final loss function that integrates the global loss term $L_g$ and local loss
term $L_l$ by weight $\lambda$ as follows:
\begin{equation}
 L = L_g + \lambda L_l
 \end{equation}
where
\begin{equation}
L_g=\sum_{j} l\left(d_{G}\left(q, p_{sh}^{q}\right)+m-d_{G}\left(q, n_{j}^{q}\right)\right)
\end{equation}
and
\begin{equation}
L_l=\sum_{j} l\left(d_{L}\left(q, p_{sh}^{q}\right)+m-d_{L}\left(q, n_{j}^{q}\right)\right).
\end{equation}

\section{Experiment}

\begin{table}[!t]
  \caption{Summary of the datasets.}
  	\vspace{-0.3cm}
  \label{tableDataset}
  \begin{center}
  \setlength{\tabcolsep}{1.3mm}{
  \begin{tabular}{cccccc}
    \toprule
    \multirow{2}{*}{Dataset} & \multicolumn{1}{c}{\multirow{2}{*}{Description}} & \multicolumn{2}{c}{Number} & \multicolumn{2}{c}{Variation} \\ \cline{3-4} \cline{5-6}
     \multicolumn{1}{c}{}& & Database & Queries & \multicolumn{1}{c}{Cond.} & \multicolumn{1}{c}{View.} \\ \cline{1-6}
    MSLS-train & long-term & 900k & 500k &  \checkmark & \checkmark \\
    MSLS-val &  urban, suburban & 19k & 11k & \checkmark & \checkmark \\ \cline{1-6}
    Pitts30k-train & urban & 10k & 7,416 & \checkmark & \checkmark \\
    Pitts30k-test & panorama & 10k & 6,816 & \checkmark & \checkmark \\ \cline{1-6}
    GP & urban, campus & 200 & 200 & \checkmark & \checkmark \\ \cline{1-6}
    Nordland & seasonal & 3,450 & 3,450  & \checkmark & $\times$ \\
    \bottomrule
  \end{tabular}}
  \vspace{-0.6cm}
\end{center}
\end{table}
\subsection{Datasets and Performance Evaluation}
Our experiments are conducted on four VPR benchmark datasets. The key information of these datasets is summarized in Table \ref{tableDataset}. Here we briefly describe the usage of these datasets.
\textbf{Mapillary Street Level Sequences (MSLS)} \cite{msls} contains over 1.6 million images captured in both urban and suburban environments over seven years. Since its test set labels have not yet been publicly available, we test the model on its validation set as in \cite{tcl,benchmark}.
\textbf{Pittsburgh (Pitts30k)} \cite{pitts} includes 30k reference images and 24k query images, and has been geographically divided into train/validation/test sets. We evaluate models on its test set.
\textbf{Gardens Point (GP)} \cite{sunderhaufIROS2015} contains images of two traverses during the day and night. We use the night-right (reference) and the day-left (query) images as test set.
\textbf{Nordland} \cite{nordland} consists of train and test set. We use the summer (reference) and winter (query) images from the down-sampled (224x224) test set in the following experiments.

We evaluate the recognition performance mainly based on Recall@N (R@N). It is the percentage of queries for which at least one of the top-N retrieved images is taken within a certain threshold of the query. The threshold is 25 meters for MSLS and Pitts30k, and $\pm 2$ frames for GP and Nordland, following standard procedure \cite{netvlad,benchmark,tcl}.

\subsection{Implementation Details}
We use the CCT-14 model pre-trained on ImageNet \cite{ImageNet} as the backbone. The 9th and subsequent transformer encoder layers are removed, and the layers before the 3rd encoder layer are frozen. All experiments are conducted on an NVIDIA GeForce RTX 3090 GPU. The input image is resized to 384×384 resolution. Adam optimizer is used to train our models with the learning rate = 0.00001 and the batch size = 4. All test models are trained on MSLS, and the test model for Pitts30k is finetuned with random horizontal flipping and random resized cropping on Pitts30k-train. The model training process stops when R@1+R@5 on the validation set does not improve for 3 epochs. (But we only use 7k queries on MSLS to train the model for the GP dataset.) An epoch is defined as passing 40k queries for MSLS and 5k queries for pitts30k.

During training, the potential positive samples are the database images within a 10 meters radius from the query, while the definite negatives are those further than 25 meters. The number of hard negative samples in triplets is 2, which are selected from 1000 random negatives. We empirically set the weight $\lambda=1$, the margin $m=0.1$. The $k$ and $k’$ in Eq. 7 equal 30\% of the number of all potential positives for MSLS, and equal 1 and 2 for Pitts30k. And we only re-rank the top-20 candidates to get the final results.

\subsection{Ablation Study}

To evaluate the effectiveness of the proposed DALF algorithm, the ShPSM strategy, as well as the joint optimization combining global loss $L_g$ and local loss $L_l$, we perform a series of ablation experiments:
\begin{itemize}
	\item\textbf {GeM(U):} Untrained vanilla GeM.
	\item\textbf {GeM(V):} Vanilla GeM trained with global loss.
	\item\textbf {GeM(J):} Use joint loss ($L_g+L_l$) of AANet for training. But only use global features for direct retrieval.
	\item\textbf {GeM(J+S):} Use joint loss ($L_g+L_l$) of AANet for training and the ShPSM strategy to mine hard positive samples. But only use global features for direct retrieval.
	\item\textbf {AANet(-S):} Use AANet for Two-stage retrieval. But do not use ShPSM to mine hard positive samples.
	\item\textbf {AANet:} Our complete pipeline.
\end{itemize}

\begin{figure}[!t]
	\centering
	\includegraphics[width=0.7\linewidth]{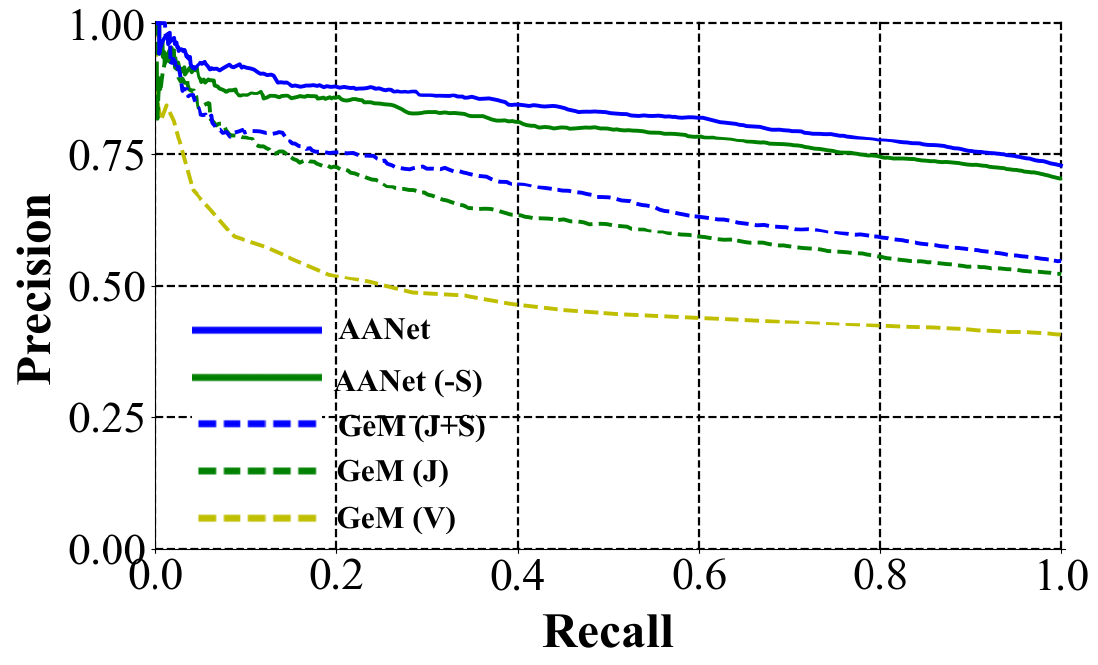}
	\vspace{-0.3cm}
	\caption{
		Precision-recall curves on the Nordland dataset.
	}
	\vspace{-0.4cm}
	\label{PR}
\end{figure}

In this subsection, we mainly use R@1 for performance evaluation. The R@1 results are shown in Table \ref{table_abl}. We also provide the PR curves on the Nordland as shown in Fig \ref{PR}. The results show that training with global loss significantly improves performance compared to off-the-shelf GeM(U). And the performance is further improved after using joint loss $L_g+L_l$ for optimization. This indicates that the combination of global loss yielded by the global (aggregation) branch and local loss yielded by the local (alignment) branch to optimize the unified backbone is successful. AANet(-S) always performs better than GeM(J), and AANet also always outperforms GeM(S), illustrating that the re-ranking method can improve performance with or without the ShPSM strategy. Likewise, GeM(J+S) always outperforms GeM(J), while AANet also always behaves better than AANet(-S). It shows that the performance can be improved by the ShPSM strategy on these datasets, with or without re-ranking. The above results indicate that the proposed DALF algorithm and ShPSM strategy work well in our AANet.

In addition, the improvements of our method on these datasets are different. The most significant improvement is achieved on the Nordland dataset, where AANet achieves more than 70\% absolute improvement over GeM(U) and more than 30\% absolute improvement over GeM(V). This is due to the extreme appearance changes caused by seasonal changing in the Nordland dataset. The network trained with hard positive samples can better cope with the appearance changes than that trained with easiest positive samples. Besides, we use patch-level local features to compute the local distance through aligning. In addition to solving the viewpoint change problem, this could also provide more detailed information (especially compared with keypoint-level local features), which is critical under severe appearance changes. Although the ShPSM strategy brings less improvement than using the DALF algorithm for re-ranking on these datasets, its process is concise and has no additional burden. It is still worthwhile to use the ShPSM strategy for training.

\begin{table}[!t]
	\caption{Results (R@1) of Different Ablated Versions.}
	\vspace{-0.3cm}	
	\label{table_abl}
	\begin{center}
		\resizebox{.92\hsize}{!}{$	
			\begin{tabular}{c|ccc|ccc}
			\toprule
			Method&\scriptsize{$L_g+L_l$}&\scriptsize{ShPSM}&\scriptsize{DALF} &MSLS  &Pitts30k &Nordland \\
			\midrule
			GeM(U) &$\times$ &$\times$ &$\times$& 20.2 & 51.6 &2.7\\
			\hline
			GeM(V) &$\times$ &$\times$ &$\times$& 72.4 & 82.5 &40.7\\
			\hline
			GeM(J) &\checkmark&$\times$&$\times$& 75.6 & 83.7 &52.3\\
			\hline
			GeM(J+S)  &\checkmark&\checkmark&$\times$& 76.5 & 84.8 &54.8\\
			\hline
			AANet(-S) &\checkmark&$\times$ &\checkmark& 79.7 & 87.2 & 70.4\\
			\hline
			AANet &\checkmark&\checkmark&\checkmark& \bf 80.1 &  \bf 88.0 & \bf 72.9\\
			\bottomrule
			\end{tabular}
			$}
	\end{center}
	\vspace{-0.4cm}
\end{table}

\subsection{Comparisons with State-of-the-Art Methods}
\begin{table*}[tbp]
\setlength\tabcolsep{6.6pt}
\begin{scriptsize}
    \begin{center}
    \caption{Comparison to State-of-the-art Methods on Four Benchmark Datasets.}
    \label{result}
    \centering
     \begin{tabularx}{17.5cm}{c|c|c|ccc|ccc|ccc|ccc}
    \toprule
       \multirow{2}{*}{Method} & \multirow{2}{*}{Dim} & Time & \multicolumn{3}{c|}{\textbf{MSLS val}} &
       \multicolumn{3}{c|}{\textbf{Pitts30k}} & \multicolumn{3}{c|}{\textbf{Nordland}}  & \multicolumn{3}{c}{\textbf{Gardens Point}} \\ 
    \cline { 4 - 15 } &&(s)& R@1 & R@5 & R@10 & R@1 & R@5 & R@10 & R @1 & R@5 & R@10 & R@1 & R@5 & R@10\\
    \midrule
    NetVLAD \cite{netvlad}  &32768 & 0.025 & 60.8 & 74.3 & 79.5 &  81.9 & 91.2 & 93.7 & 7.7 & 13.7 &17.7 & 38.0& 74.5& 83.5 \\
    GCL \cite{gcl} & 2048 & 0.061  & \textbf{78.3} & 85.0 & 86.9 & 75.6 & 91.5 & 92.6 & 27.2 & 45.4 & 53.7 & 42.0 & 79.5 & 88.0 \\ % 
    \textbf{AANet (w/o re-ranking)}& 384 & \textbf{0.014} & 76.5 & \textbf{87.4} & \textbf{90.3} & \textbf{84.8} & \textbf{93.4} & \textbf{95.5} & \textbf{54.8}  & \textbf{77.9} & \textbf{85.7} & \textbf{54.0} & \textbf{89.0} & \textbf{93.5} \\
    \midrule
    SP-SuperGlue \cite{sp,sg} & 256 & - & 78.1  & 81.9 & 84.3 & 87.2 &  94.8 & 96.4  & 29.1 & 33.5 & 34.3 & - & - & -  \\
    Patch-NetVLAD-s \cite{patchvlad} & 512 & 0.90 & 77.0 & 84.2 & 86.9 & 87.1 & 94.0 & 95.6 & 34.9 & 49.8 & 53.3 & 71.5 & 93.5 & 96.5  \\
    Patch-NetVLAD-p \cite{patchvlad} & 4096 & 16.51 & 79.5 & 86.2 & 87.7 & 88.7 & 94.5 & 95.9 & 46.4 &  58.0 &60.4 &78.0 & 94.0 & 97.0 \\
    TCL \cite{tcl} & 384 & - & 78.7 & 82.5 & 85.3 & \textbf{90.5}& \textbf{95.9}& \textbf{97.5} & 43.5 & 51.7 & 54.3 & \textbf{80.5} & 95.0 & 97.0 \\
    \textbf{AANet (ours)}& 384 & \textbf{0.053} & \textbf{80.1} & \textbf{88.9} & \textbf{91.0} & 88.0 & 94.2 & 95.8 & \textbf{72.9} & \textbf{87.6} & \textbf{90.6} &77.0 &\textbf{97.0} &\textbf{99.5} \\
    \bottomrule
  \end{tabularx}
\end{center}
\end{scriptsize}
\vspace{-0.5cm}
\end{table*}

We evaluate the recognition performance by comparing AANet against several state-of-the-art VPR methods, including two VPR methods using global features for direct retrieval: NetVLAD \cite{netvlad} and GCL \cite{gcl}, and three two-stage VPR methods: SP-SuperGlue \cite{sp,sg}, Patch-NetVLAD \cite{patchvlad} and TCL \cite{tcl}. The NetVLAD was trained on the Pitts250k. GCL (ResNet152-GeM-PCA) and TCL (DeiT-S, TCL-R100) follow the
optimal testing configuration of the original work. For Patch-NetVLAD, our approach is compared with its two different versions, where Patch-NetVLAD-s is speed-focused and Patch-NetVLAD-p is performance-focused. SP-SuperGlue was first used for VPR in \cite{patchvlad}. It retrieved the candidates by NetVLAD, then used the SuperGlue \cite{sg} matcher to match the SuperPoint \cite{sp} local descriptors for re-ranking. The quantitative results are shown in Table \ref{result}. 

Our method is denoted as AANet (w/o re-ranking) when it does not use re-ranking. Although we use 384-dimensional global features, which are more compact than those of NetVLAD and GCL, it performs better on all four datasets (except R@1 on MSLS) in comparison with these two methods. When compared with GCL, it has achieved 2.4\%, 1.9\%, 32.5\%, and 9.5\% absolute improvements for R@5 on the MSLS, Pitts30k, Nordland, and GP datasets, respectively. 

AANet also achieves competitive results in comparison with two-stage methods that require re-ranking. It outperforms other methods on MSLS, Nordland, and GP (except R@1). The R@1 of AANet on the Pitts30k dataset is better compared to SP-SuperGlue and Patch-NetVLAD-s, but worse than Patch-NetVLAD-p, which used RANSAC for spatial verification. We think this is because the place images of Pitts30k have a large rotational variation. The DALF algorithm in AANet mainly considers horizontal and vertical shifts. It is not good at addressing rotational changes. Similar to this, Rapid Spatial Scoring used in Patch-NetVLAD-s also considers horizontal and vertical movements. But our method outperforms it on all datasets. Benefiting from our re-ranking method, more than absolute improvements of 18\% for R@1 are achieved on both Nordland and GP datasets compared to the case without re-ranking.

\subsection{Runtime Analysis}
We analyze the computational efficiency of the VPR methods by the running time (feature encoding time + feature matching time) of each query on the Pitts30k test dataset. We compare the retrieval time of the first stage of our method with two direct retrieval approaches (NetVLAD and GCL) and the overall time consumption with Patch-NetVLAD (a two-stage method). The results are also shown in Table \ref{result}. Since our method uses 384-D global features in the first stage, whose size is much smaller than that used in NetVLAD and GCL, it has a significant advantage in computational efficiency (it is also due to the different backbones). Moreover, our method is more than two orders of magnitude faster than Patch-NetVLAD-p, which uses RANSAC for geometrical consistency verification. Although a fast spatial verification method (Rapid Spatial Scoring) is proposed in Patch-NetVLAD-s, it is still more time-consuming than our method. It is worth mentioning that to extend the horizontal regional features alignment in STA-VPR \cite{sta-vpr} and TCL \cite{tcl} to both horizontal and vertical directions, a naive solution is to first align the local features in all pairs of regional features in the vertical direction using the DTW-based algorithm (requiring $N\times N$ times alignments), and then align regional features in the horizontal direction. In this way, a total of $N\times N+1$ times DTW-based alignments are performed. However, our proposed DALF method only needs to perform the DTW-based alignment process twice, which greatly improves the alignment efficiency.

\section{Conclusion}
Here we proposed the hierarchical VPR architecture AANet that utilizes the aggregation module to extract global features for retrieving candidate places and the DALF alignment module to align local features under spatial constraints for re-ranking. Compared to the mainstream re-ranking methods that require geometric consistency check, our method is significantly more efficient. To overcome the limitation that many methods use the easiest positive samples during training and have difficulty in identifying harder positive pairs during inference, we proposed the ShPSM strategy to mine semi-hard positive images for training a more robust network. The experimental results illustrated that our method outperformed the state-of-the-art methods on several benchmark datasets while showing high computational efficiency.
\nocite{*}
\bibliographystyle{IEEEtran}
\bibliography{aanet}

\end{document}